\documentclass[lettersize,journal]{IEEEtran}
\usepackage{amsmath,amsfonts}
\usepackage{algorithmic}
\usepackage{algorithm}
\usepackage{array}
\usepackage[caption=false,font=normalsize,labelfont=sf,textfont=sf]{subfig}
\usepackage{textcomp}
\usepackage{stfloats}
\usepackage{url}
\usepackage{verbatim}
\usepackage{graphicx}
\usepackage{cite}
\usepackage{multirow}
\begin{document}

\title{Domain Adaptation based on Human Feedback for Enhancing Generative Model Denoising Abilities}

\author{Hyun-Cheol Park, Sung Ho Kang
        % <-this % stops a space
\thanks{\textit{Hyun-Cheol Park is with Division of Industrial Mathematics, National Institute for Mathematical Sciences, South Korea. \\ E-mail: hc.park@nims.re.kr}}% <-this % stops a space
\thanks{\textit{Sung Ho Kang is with Division of Industrial Mathematics, National Institute for Mathematical Sciences, South Korea. \\ E-mail: runits@nims.re.kr}}
\thanks{\textit{Sung Ho Kang is a correspondence author}}}
\markboth{Journal of \LaTeX\ Class Files,~Vol.~14, No.~8, August~2021}%
{Shell \MakeLowercase{\textit{et al.}}: A Sample Article Using IEEEtran.cls for IEEE Journals}

\IEEEpubid{0000--0000/00\$00.00~\copyright~2021 IEEE}

\maketitle

\begin{abstract}
How can we apply human feedback into generative model? As answer of this question, in this paper, we show the method applied on denoising problem and domain adaptation using human feedback. Deep generative models have demonstrated impressive results in image denoising. However, current image denoising models often produce inappropriate results when applied to domains different from the ones they were trained on. If there are `Good' and `Bad' result for unseen data, how to raise up quality of `Bad' result. Most methods use an approach based on generalization of model. However, these methods require target image for training or adapting unseen domain. In this paper, to adapting domain, we deal with non-target image for unseen domain, and improve specific failed image. To address this, we propose a method for fine-tuning inappropriate results generated in a different domain by utilizing human feedback. First, we train a generator to denoise images using only the noisy MNIST digit '0' images. The denoising generator trained on the source domain leads to unintended results when applied to target domain images. To achieve domain adaptation, we construct a noise-image denoising generated image data set and train a reward model predict human feedback. Finally, we fine-tune the generator on the different domain using the reward model with auxiliary loss function, aiming to transfer denoising capabilities to target domain. Our approach demonstrates the potential to efficiently fine-tune a generator trained on one domain using human feedback from another domain, thereby enhancing denoising abilities in different domains.
\end{abstract}


\section{Introduction}
\IEEEPARstart{D}{eep} generative models have achieved remarkable success in image generation tasks \cite{goodfellow2020generative, DonahueKD17, mirza2014conditional}. In particular, generative adversarial networks (GANs) are widely known as a fundamental theory that demonstrates how to generate realistic images. Recently, GANs are also utilized for specific purposes such as image denoising \cite{zhang2017beyond, tran2020gan, vo2021hi}, super-resolution \cite{haris2018deep, wang2018esrgan}, and style transfer \cite{zhu2017unpaired, zhu2017toward, yi2017dualgan, liu2017unsupervised}. These objectives involve training GANs using supervised learning with paired data sets aiming to learn the target distribution, which has been shown to yield successful results.

However, despite the impressive performance of these models within their training domain, they often encounter challenges when applied to unseen domains, resulting in subpar outputs. In the context of GANs based on image-to-image generation \cite{tian2020deep}, which aim to preserve the original intrinsic characteristics while learning the target distribution, both successful and unsuccessful cases can emerge during testing on unseen domains. For example, when presented with ten samples from an unseen domain, seven of them may yield successful translations, while the remaining three produce unsatisfactory results. This raises the question: should we simply discard these three failed samples, or is there a way to enhance and improve them to achieve better outcomes? Obtaining ground-truth data for the unseen domain could facilitate domain-specific training; however, in many practical scenarios, acquiring target data for unseen domains poses significant challenges. As an alternative, applying domain adaptation methods \cite{volpi2018adversarial, wang2018transferring, kang2019contrastive, alanov2022hyperdomainnet, bousmalis2017unsupervised, lin2019real, chen2022reusing, kwon2023one} can mitigate this issue, but even domain-adapted models may still yield failed results based on human preferences.

\begin{figure}[!t]
\centering
\includegraphics[width=3.3in]{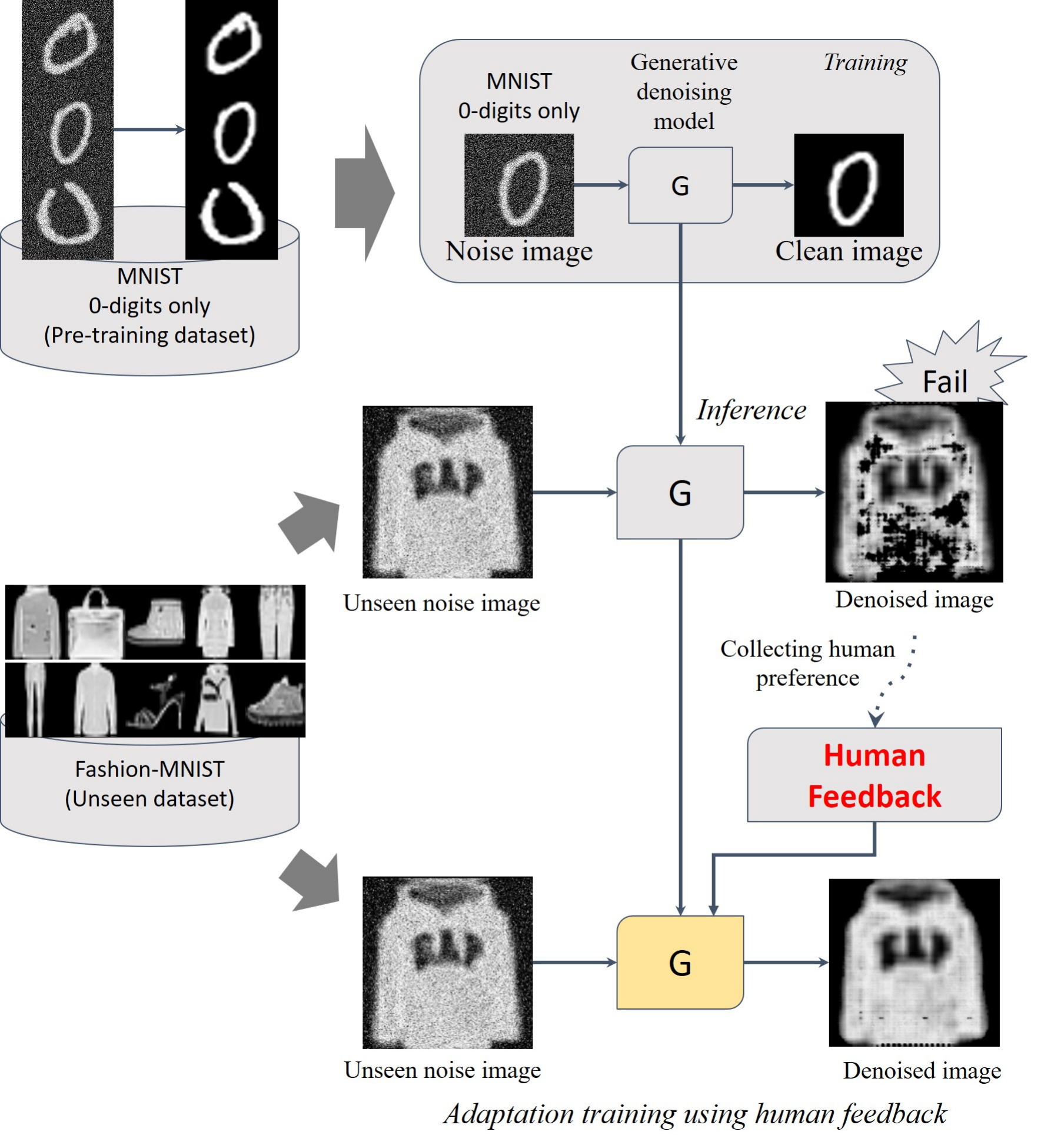}
\caption{Overview for adaptation training based on human feedback.}
\label{fig_1}
\end{figure}
\IEEEpubidadjcol
Our objective diverges from conventional domain adaptation approaches. As demonstrated in \cite{volpi2018adversarial,lin2019real}, domain adaptation focuses on training methods that aim to minimize the distinguishability between the source and target domain distributions in the latent space. This macro-level approach seeks to minimize the overall gap between source and target domains. However, at a micro-level, there remain opportunities for improvement in the generated results. Therefore, our goal is to address and rectify instances of failure within the output produced by the trained model.

Similarly, during the training of ChatGPT\cite{brown2020language}, it excels at generating high-quality language responses through extensive pre-training on vast data sets. Nevertheless, upon human evaluation, the generated sentences may exhibit a dichotomy: some appear naturally flowing, while others seem less fluent. To bridge this gap, ChatGPT\cite{brown2020language} leverages Human-Feedback\cite{NIPS2017_hf} to enhance its ability to produce more seamlessly natural sentences. Furthermore, in the domain of aligning text-to-image models\cite{lee2023aligning}, the introduction of human feedback has demonstrated significant improvements in model performance. However, research on model refinement through human feedback in GANs is still scarce, and through our paper, we aim to showcase the potential of model refinement through the profound influence of human feedback.

Recently, drawing inspiration from the success of reinforcement learning human feedback (RLHF) \cite{NEURIPS2020_learning_summary_hf} in language domains, we present an innovative approach for unseen domain adaptation based on human feedback. Analogous to how children learn from the feedback provided by their parents, we adopt a similar strategy. For instance, if a child learns how to remove noise from the background of a single image, they can subsequently apply denoising techniques to new images. While the quality of the denoised image may vary, receiving feedback from a parent can lead to improvement. Even if we cannot surpass our previous achievements, we can still imitate and learn from them. This approach shows promise in addressing the challenges of unsupervised unseen domain adaptation and opens new possibilities for model enhancement through the profound influence of human feedback. As we explore this innovative avenue, our aim is to make significant contributions to the field of AI and foster advancements in unsupervised domain adaptation research. The guiding philosophy behind our work can be summarized as: ``Our goal is to learn what I am not good at, just like what I am good at."

To achieve this, we introduce a deep feedback network that utilizes human feedback to adaptation unlabeled target domain. To replicate restricted learning circumstances, we conduct experiments in the denoising problem. Initially, we train the model using a restrictive training approach, focusing solely on denoising the digit `0' within the MNIST data set. Subsequently, we evaluate the model's performance on the Fashion-MNIST data set, which represents an unseen domain. It becomes evident that the pre-trained model, trained on MNIST, produces unintended results when applied to the unseen domain. To adapt to the unseen domain, we introduce a training method based on human feedback. Human feedback assesses the model's results in the unseen domain as either `Good' or `Bad'. The model is then fine-tuned using the gradient of these assessments. This approach shows promising potential for efficiently fine-tuning the model using feedback from generators trained on other domains.

\subsubsection*{\bf We can summarize our main contributions as follows}
\begin{itemize}
\item{We propose adaptation method for the domain of image-based generative models through human feedback.}
\item{We perform domain adaptation while maintaining the quality of the generated image using an optional loss function with a reward model using the human feedback.}
\item{We show that the model can be adapted by human feedback, even in the absence of labeled target data}
\end{itemize}

\section{Methods}
Our overall process consists of three steps. First, the denoising model is pre-trained in the basis domain, serving as the fundamental ability for denoising. Next, the reward model is trained using human feedback. To train the reward model, humans manually annotate denoised images as either $Good$ or $Bad$. Finally, the basis generator is re-trained using the reward model. Even if the generator produces denoised images of low quality, it will be trained to prioritize good results based on the provided human feedback.

\subsection{Pre-training basis domain for denoising}
In this step, we focused on creating an intentional class-biased generator. The model is trained to acquire the fundamental ability of denoising using simple images as shown in Step 1 of Fig. \ref{training_step}. The architecture of the model consists of generative adversarial networks (GANs). We employed the pix2pix\cite{isola2017image} model as our baseline, which relies on paired training.  To train the model, a paired data set is required, consisting of both clean and noise images. For our paired training data set, we used the only 0-digit in MNIST data set. To create a pair, selected 0-digit images and combined with synthesized noise. Consider the synthesized noise image $z$, which is a 2D image represented as $z \in R^{m\times n}$. It is composed of both the original image and noise, denoted as $x$ and $n$, respectively:
\begin{equation}
\label{deqn_ex1a}
z = x+n.
\end{equation}
We assume that the clean image is selected from the source domain. Therefore, the synthesized noise image $z$ and the original image $x$ are treated as paired data. For convenience notation, source and unseen domain data denote as $z_s$ and $z_u$, respectively.

The generator is trained to produce samples of good quality from input noise variables $p_n$. To train the model on the source domain, the final loss is defined as follows:
\begin{equation}
\label{pre_training_final_loss}
L_{step_1}(G_s,D) = L_{GAN}(G_s,D) + L_{pixel-wise}(G_s)
\end{equation}
where the samples $G_s(z_s)$ obtained when $z_s \sim p_n$ follow a distribution that represents good quality in source domain. In other words, The generator $G_s$ is trained to learn the mapping from the noise image $z$ to the clean image $x$, denoted as $G_s:z_s \rightarrow x$. The objective of the generator is to estimate the distribution of $x$, denoted as $G_s(z_s) \approx x$. To achieve this, the GAN consists of an adversarial discriminator $D$, which distinguishes between `Real' and `Fake' images. `Real' refers to the original image $x$, while 'Fake' corresponds to the generated image $G_s(z_s)$ produced by the generator. Both the generator $G_s$ and the discriminator $D$ are trained adversarially. The objective function can be expressed as follows:
\begin{multline}
\label{basic_gan_loss}
\underset {G_s} {\text{min}} \ \underset {D} {\text{max}} \ L_{GAN}(G_s,D) = \mathbb{E_{\text{$z_s, x$}}} \ [\log{D(z_s,x)}] + \\ \mathbb{E_{\text{$z_s\sim p_n(z_s)$}}} \ [\log{(1-D(z_s,G_s(z_s)))}].
\end{multline}
where $G_s(z_s)$ represents the generation of a clean image from a noise image $z_s$. The discriminator $D$ is responsible for classifying between the real and fake distributions. In order to induce a mistake in $D, G_s$ aims to minimize Equation (\ref{basic_gan_loss}). On the other hand, $D$ maximizes the objective function to distinguish between real and generated images. 

In our study, we tackle the problem of denoising while preserving the underlying morphological structures. Traditional GAN \cite{goodfellow2020generative} frameworks approximate the target distribution during training. However, in the context of image processing, the generated images may inadvertently alter the essential morphological characteristics of the originals\cite{isola2017image,zhang2019noise,park2022unpaired}. To mitigate this issue and ensure the preservation of morphological structures, an auxiliary loss term is incorporated into the objective function:
\begin{equation}
\label{L_pixel-wise}
L_{pixel-wise}(G_s) = \mathbb{E_{\text{$z_s,x$}}} \ [\Vert{x-G_s(z_s)}{\Vert}_1].
\end{equation}
Similar to the \cite{isola2017image} approach, the auxiliary loss employs the $L1$ distance between the target image $x$ and the generated image $G_s(z_s)$. To train model on source domain, final loss is as follows:

\begin{figure*}[!t]
\centering
\includegraphics[width=7in]{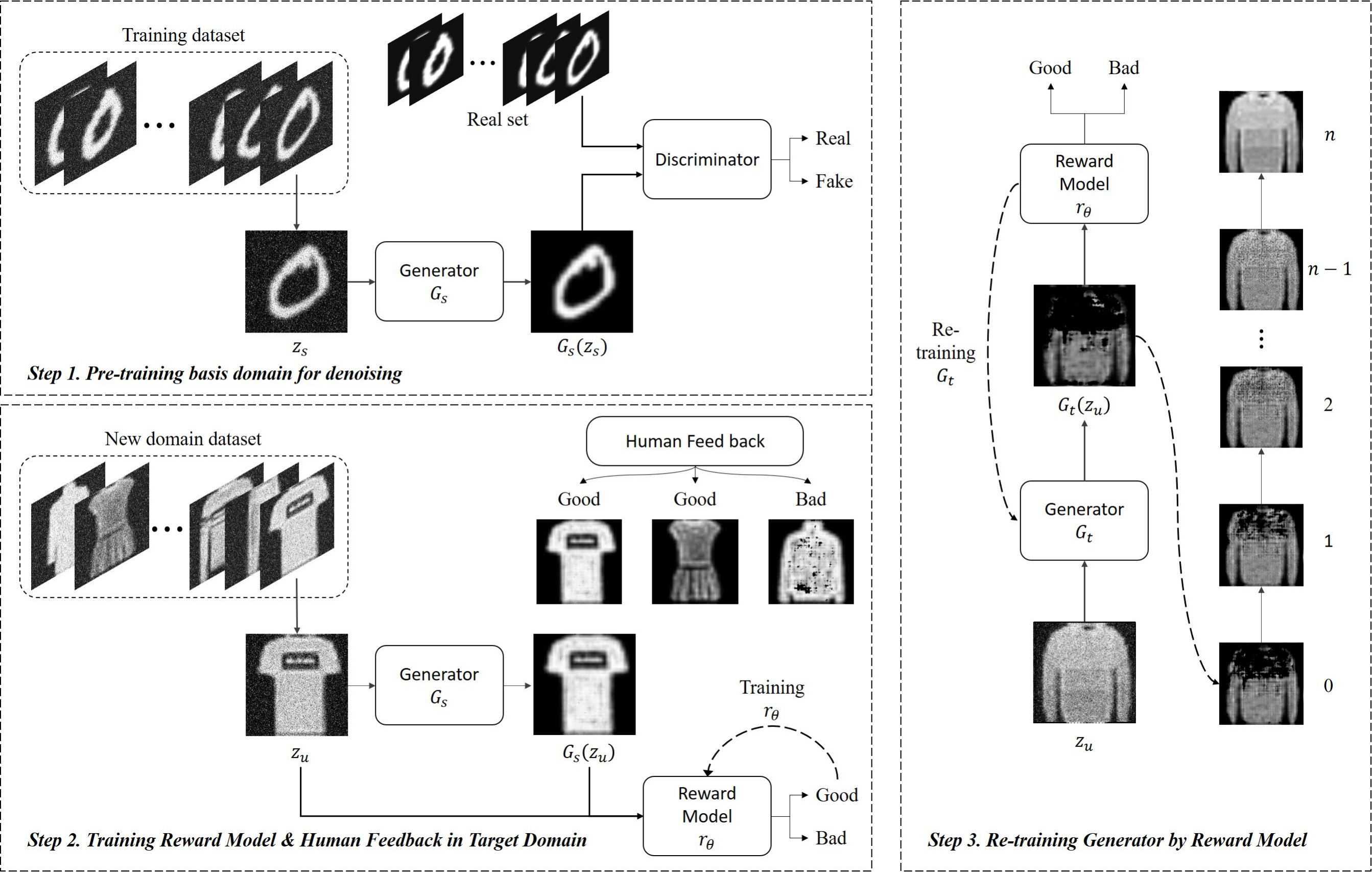}
\caption{The training data set `0' is sourced from the MNIST data set, while the new domain data set is the Fashion-MNIST data set. The $G_s$ model in Step 2 is trained on the MNIST data set during Step 1. Subsequently, the $G_s$ model in Step 3 is fine-tuned using the reward model based on human feedback.}
\label{training_step}
\end{figure*}

\subsection{Human feedback and training reward model}
The integration of human feedback has demonstrated high adaptability across various domains \cite{lee2023aligning,ouyang2022training,stiennon2020learning}. This valuable information is used to train a reward model, which acts as a substitute for human assessment and enhances the model's performance. In this section, we provide a detailed description of how human feedback is gathered. In the previous section, we presented the basic denoising GAN model using pix2pix\cite{isola2017image}, which we referred to as the Supervised Denoising Model (SDM). Human feedback is obtained through manual assessments of the SDM results from unseen domain samples $z_u$. The assessments are categorized as `Good' if the image was clean and `Bad' if the image contained noise or collapse. (see Step 2 in Fig. \ref{training_step}.)

The assessments `Good' and `Bad' are utilized as ground truth labels ($y_r={0,1}$) to train the reward model. The reward model, denoted as $r_\theta$, follows the same architecture as the discriminator in the SDM. The loss function for $r_\theta$ is as follows:
\begin{multline}
\label{reward_loss}
L_{reward}(\hat{G_s}, r_\theta) = \underset {r_\theta} {\text{min}} \mathbb{E}_{z_u\sim p_n(z_u)}-[y_r \log{r_\theta(\hat{G_s}(z_u), z_u)}
\\+(1-y_r)\log{(1-r_\theta(\hat{G_s}(z_u),z_u)})].
\end{multline}
where $\hat{G_s}$ is a frozen denoising generator model using source domain. During the training of the reward model, $\hat{G_s}$ remains untrainable and is solely used to generated denoised images. $r_\theta$ assesses these denoised images and is trained using $y_r$ labels. Notably, the reward model can be trained to capture human preferences, as the $y_r$ labels are collected through human feedback.

\subsection{Objective}
In this section, we present the final formulation of the loss function, which consists of auxiliary terms. Each auxiliary term includes reward loss, consistency loss, and regularization loss, used to train $G_t$. Here, $G_t$ represents the adapted model which is fine-tuned from $G_s$ in the unseen domain. Thus, the architecture and initial parameters of $G_t$ are the same as those of $G_s$. \\

\noindent {\bf{Reward Loss $L_r$:}} The primary objective of Generator $G_t$ is to generate denoised images that are assessed by the reward model as `$Good$'(0, indicating clean images). The minimization of $L_r$ aims to train the generator $G_t$ to generate clean images. In other words, the reward loss $L_r$ trains $G_t$ to map from the distribution of `$Bad$' quality images (distribution $j$) to the distribution of `$Good$' quality images (distribution $k$), $G_t: j\rightarrow k$.
\begin{equation}
\label{Lr_loss}
L_{r}(G_t) = \mathbb{E}_{z_u\sim p_n(z_u)} \ [-\log{(1-\hat{r_\theta}(G_t(z_u), z_u)}].
\end{equation}
where $\hat{r_\theta}$ is a reward model trained on human feedback and has fixed parameters. Thus, $\hat{r_\theta}$ only assesses the quality of the generated image from $G_t$ and the input image $z_u$.

In this context, by fine-tuning $G_t$ from $G_s$ using $L_r$ loss, $G_t$ is able to closely approximate the $x\sim p_{data}$ distribution represented by $r_\theta$ in an unseen domain. However, relying solely on $L_r$ loss for training $G_t$ may lead to over-fitting and the risk of distorting the morphological information of the original images. To alleviate this problem, we describe `Regularization Loss' and `Consistency Loss' as follows.\\

\noindent {\bf{Consistency Loss $L_p$:}} As the model learns from new data, there is a potential issue of the performance of past good results deteriorating due to parameter updates. This is commonly referred to as the problem of catastrophic forgetting. To control this issue, it is necessary to compare the outcomes of the initial parameters with the current results. We present a novel compensatory term, denoted as $L_p$, which facilitates a comparison between the outputs of the initial frozen generator, $\hat{G_s}$, and the target generator $G_t$. The primary objective of $L_p$ is to minimize the pixel-wise $L_1$ loss between the outcomes generated by $\hat{G_s}$ and $G_t$, thereby ensuring that the current model preserves crucial insights acquired from the initial generator throughout the training procedure. By incorporating this approach, we effectively address the issue of neglecting important details and consequently witness a notable enhancement in the overall performance of the current model.
\begin{multline}
\label{Lp_loss}
L_{p}(G_t) = \\
\mathbb{E}_{z_u\sim p_n(z_u)} \ [\sigma(\hat{r_\theta}(\hat{G_s}(z_u), z_u))\Vert \hat{G_s}(z_u)-G_t(z_u) {\Vert}_1].    
\end{multline}

\begin{equation}
\label{sigmoid}
\sigma(r) = \begin{cases} 0   & \text{if } r \geq \epsilon \\
        1   & \text{if } r < \epsilon
    \end {cases}
\end{equation}
where $\sigma$ is the step function, and $r$ denotes the result of the reward. $\epsilon$ is a threshold value ranging from 0 to 1.\\

\begin{figure}[!t]
\centering
\includegraphics[width=3.4in]{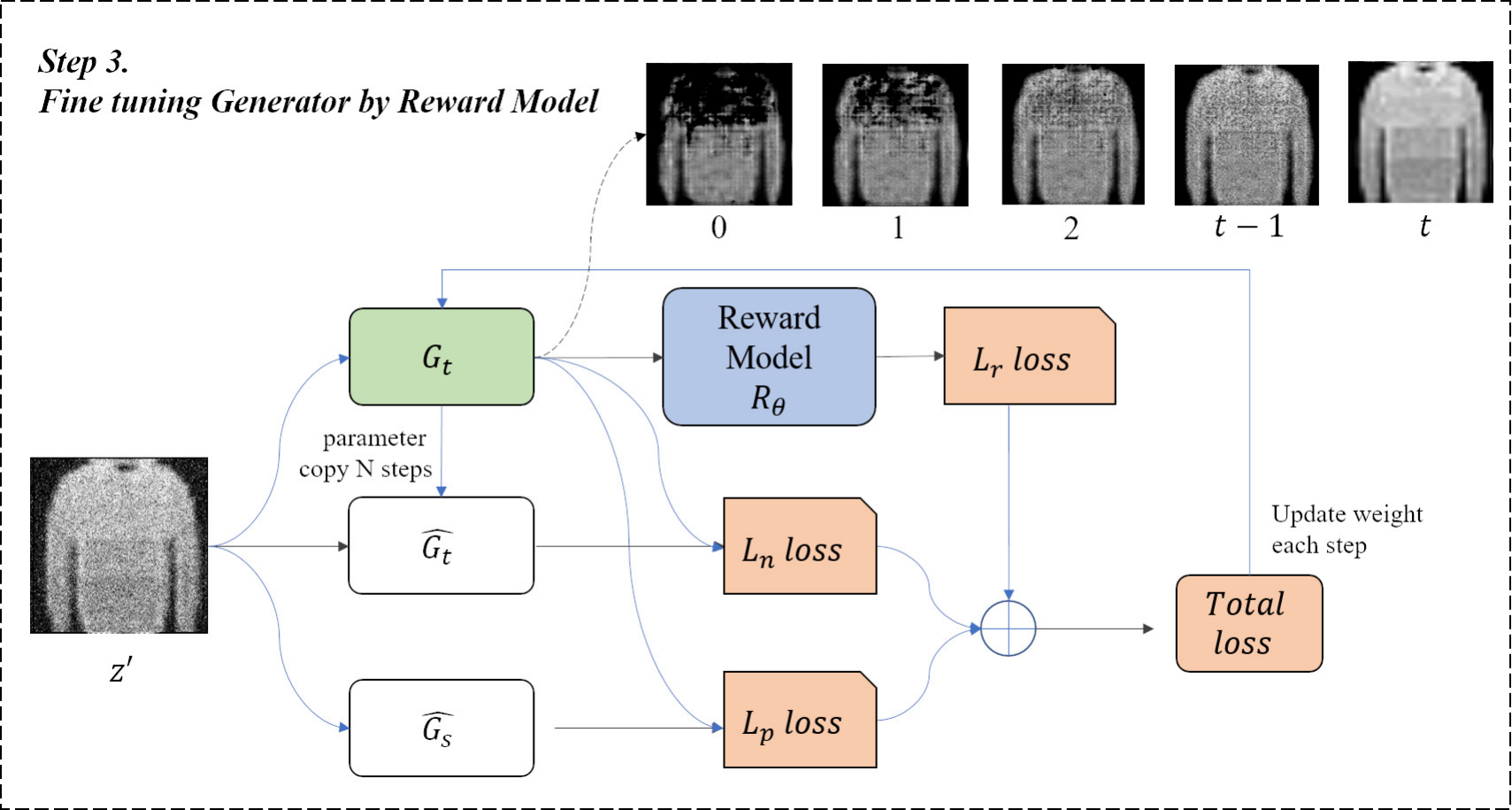}
\caption{Flow diagram of the fine tuning final objective loss functions with each objective loss }
\label{step3_total_loss}
\end{figure}

\noindent {\bf{Regularization Loss $L_n$:}} 
We employ a regularization loss term to address the issues of over-fitting and mode collapse. In existing methods, the difference in cosine similarity of feature vectors in the latent space has been compared \cite{zhu2021mind, kwon2023one}. However, in our approach, we intuitively compare the outputs of the model from the past and the current training stages to suppress excessive variations caused by the model's learning. $L_{n}$ calculates the pixel-wise $L_1$ loss between the results of the current generator and the ($n-i$)$th$ generator $\hat{G_t}$. The generator $\hat{G_t}$ copies weights from $G_t$ every $N$ steps and then freezes them.
\begin{equation}
\label{Ln_loss}
L_{n}(G_t) = \mathbb{E}_{z_u\sim p_n(z_u)} \ [\Vert \hat{G_t}(z_u)-G_t(z_u) {\Vert}_1].
\end{equation}

The final loss used to train $G_t$ is as follows:
\begin{equation}
\label{final_loss}
\underset {G_t} {\text{min}} L(G_t) = L_r(G_t)+ \alpha L_p(G_t) + (1-\alpha) L_{n}(G_t).
\end{equation}
where, $\alpha$ is control the relative balance of the two $L_p$ and $L_n$ losses. In the ablation study, we analyze the impact of each auxiliary loss on the final loss. Fig. \ref{step3_total_loss} represents the flow diagram of the more specific final objective loss functions we designed.

\section{Experiments}

\subsection{Data sets}
We utilized two data sets in our experiments: MNIST \cite{lecun1998gradient} and Fashion-MNIST \cite{xiao2017fashion}. MNIST is a grayscale image data set consisting of 10 classes representing digits from 0 to 9. Each MNIST image has dimensions of 28$\times$28 pixels. The data set comprises a training set of 60,000 images and a test set of 10,000 images.

In the experiments, the MNIST data set serves as the source domain for training the initial denoising generator. Specifically, only the `0' digit is used for restrictive training on the source domain. The training set consists of 6,000 samples, and the validation set contains 1,000 samples. The MNIST images are resized to a size of 256$\times$256 pixels using bicubic interpolation. To train the initial denoising generator, a pair of data is required, consisting of clean and noisy images. The original MNIST data set is used as the clean image counterpart, while the noisy images are created by introducing artifact noise in the form of salt and pepper noise with Gaussian noise. Proportion of salt and pepper noise is equal amounts 0.5 and Gasussian noise is mean of zero and a standard deviation is 0.05.

Fashion-MNIST is a data set consisting of images representing 10 types of fashion items. It also includes a training set of 60,000 images and a test set of 10,000 images. Fashion-MNIST is employed to evaluate the model's performance and train adaptive learning. The images in Fashion-MNIST are resized to 256$\times$256 pixels and similarly augmented with noise, as done with the MNIST data set. 

\begin{table*}[t] 
\caption{Result of fine tuning using human feedback. Each row corresponds to the outcomes under different conditions of the loss function. The first row represents our proposed results. The second row shows results without $L_p$ loss, the third row shows results without $L_n$ loss, and the fourth row shows results using only $L_r$ loss. The fifth row presents results from the model trained on the source domain, and the last row displays the baseline results between noisy and clean images.}
\label{result_FT_HF}
% \resizebox{\textwidth}{!}{
% \begin{tabular}{rccccccc}
\begin{center}
\begin{tabular}{>{\centering}m{3.2cm}>{\centering}m{2cm}>{\centering}m{2cm}>{\centering}m{2cm}>{\centering}m{2cm}>{\centering}m{2cm}m{2cm}>{\centering}m{2cm}m{2cm}} 
\hline
\multirow{2}*{} & \multicolumn{2}{c}{MNIST test (10k)} & \multicolumn{2}{c}{Fashion-MNIST test (10k)} & \multicolumn{2}{c}{Fashion-MNIST train (60k)} \\
\cline{2-7}
& PSNR & SSIM & PSNR & SSIM & PSNR & SSIM\\
\hline
\textbf{$\boldsymbol{G_t(z)}$ vs x} & \textbf{29.36$\pm$0.92} & \textbf{0.95$\pm$0.01} & \textbf{25.68$\pm$3.91} & \textbf{0.84$\pm$0.11} & \textbf{25.75$\pm$3.86} & \textbf{0.84$\pm$0.10}\\
$G_t(z)$ vs $x$ / wo $L_p$ loss & 24.20$\pm$0.65 & 0.66$\pm$0.08 & 25.00$\pm$2.44 & 0.68$\pm$0.10 & 25.07$\pm$2.37 & 0.69$\pm$0.10\\
$G_t(z)$ vs $x$ / wo $L_n$ loss & 29.10$\pm$0.98 & 0.95$\pm$0.01 & 25.30$\pm$4.35 & 0.83$\pm$0.12 & 25.41$\pm$4.25 & 0.83$\pm$0.12\\
$G_t(z)$ vs $x$ /\; only $L_r$ loss & 20.66$\pm$0.69 & 0.82$\pm$0.03 & 17.97$\pm$2.15 & 0.58$\pm$0.13 & 18.03$\pm$2.14 & 0.58$\pm$0.12\\
$G_s(z)$ vs $x$  & 29.26$\pm$1.04 & 0.94$\pm$0.01 & 24.18$\pm$5.57 & 0.80$\pm$0.16 & 24.27$\pm$5.52 & 0.80$\pm$0.16\\
Baseline source ($z$ vs $x$) & 14.72$\pm$0.06 & 0.12$\pm$0.01 & 13.23$\pm$0.13 & 0.07$\pm$0.13 & 13.23$\pm$0.13 & 0.07$\pm$0.02\\
\hline
\end{tabular}
\end{center}
% }
\end{table*}

\subsection{Training setting}
\noindent {\bf{Pre-training for denoising:}} ``pix2pix"\cite{isola2017image} is employed as the baseline model in this experiment. The main objective of most GANs is to establish a mapping $G: Z\rightarrow X$. ``pix2pix" demonstrated the training approach for pixel-wise mapping between input and output images. Consequently, the generator of ``pix2pix" can effectively learn the transformation from the noise space $Z$ to the clean space $X$. In this experiment, we trained a denoising model, denoted as $G_s$, using the MNIST data set. $G_s$ was specifically trained using a set of 1,000 image pairs consisting of clean digits and their corresponding noisy versions. The clean images used in the training process were specifically selected to represent the digit `0'. For optimization, we employed the Adam solver \cite{kingma2014adam} with a batch size of 10, a learning rate of 0.0002, and momentum parameters $\beta_1=0.5$ and $\beta_2=0.999$. The denoising model was trained for 200 epochs.\\

\noindent {\bf{Inference and human feedback:}} In this paper, our proposed method demonstrates the adaptability of a pre-trained model to a target domain through human feedback. To gather human feedback, the pre-trained generate model $G_s$ is used to infer results in the target domain, which are then manually assessed by human evaluators. In our experiments, we employ Fashion-MNIST as the target domain data set, and we collect human feedback for the 10,000 test images in this data set.\\

\noindent {\bf{Training for reward model by human feedback:}} The reward model, denoted as $r_\theta$, is utilized in the auxiliary loss term. The architecture of the reward model is designed to be the same as the discriminator of the ``pix2pix" model. The hyperparameters used for training $r_\theta$ remain consistent with the pre-training setting.\\

\begin{figure}[!t]
\centering
\includegraphics[width=3.4in]{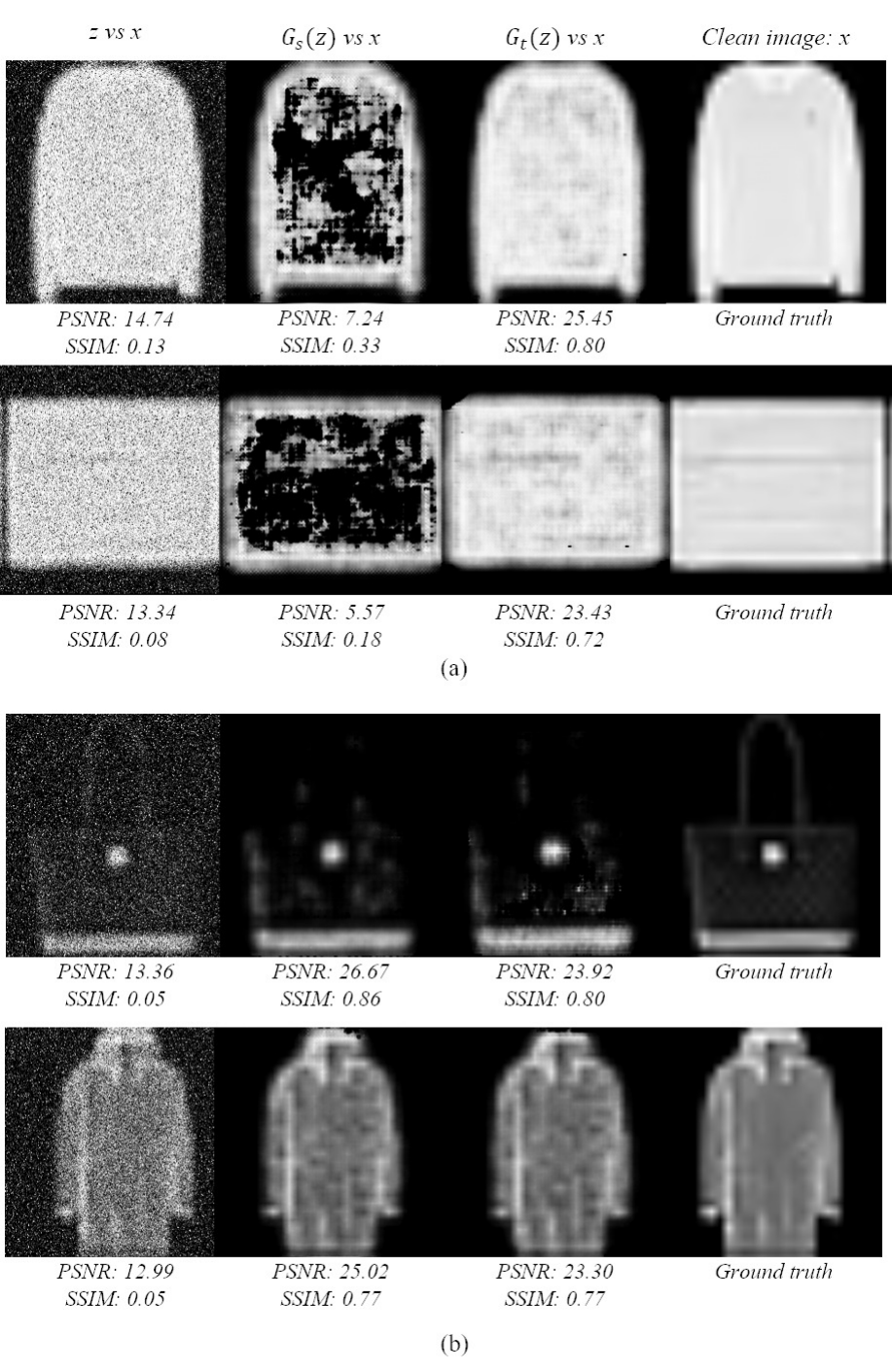}
\caption{Visual results for adaptation. The PSNR and SSIM values for each image are calculated with respect to the ground truth. $G_s$ represents the model pre-trained on MNIST, while $G_t$ represents the model fin-tuned from $G_s$ using human feedback. (a) Sample images with the most significant increase in PSNR from $G_s$ and $G_t$ output. (b) most decreasd PSNR images }
\label{visual_result_for_denoising}
\end{figure}

\begin{figure}[!t]
\centering
\includegraphics[width=3.4in]{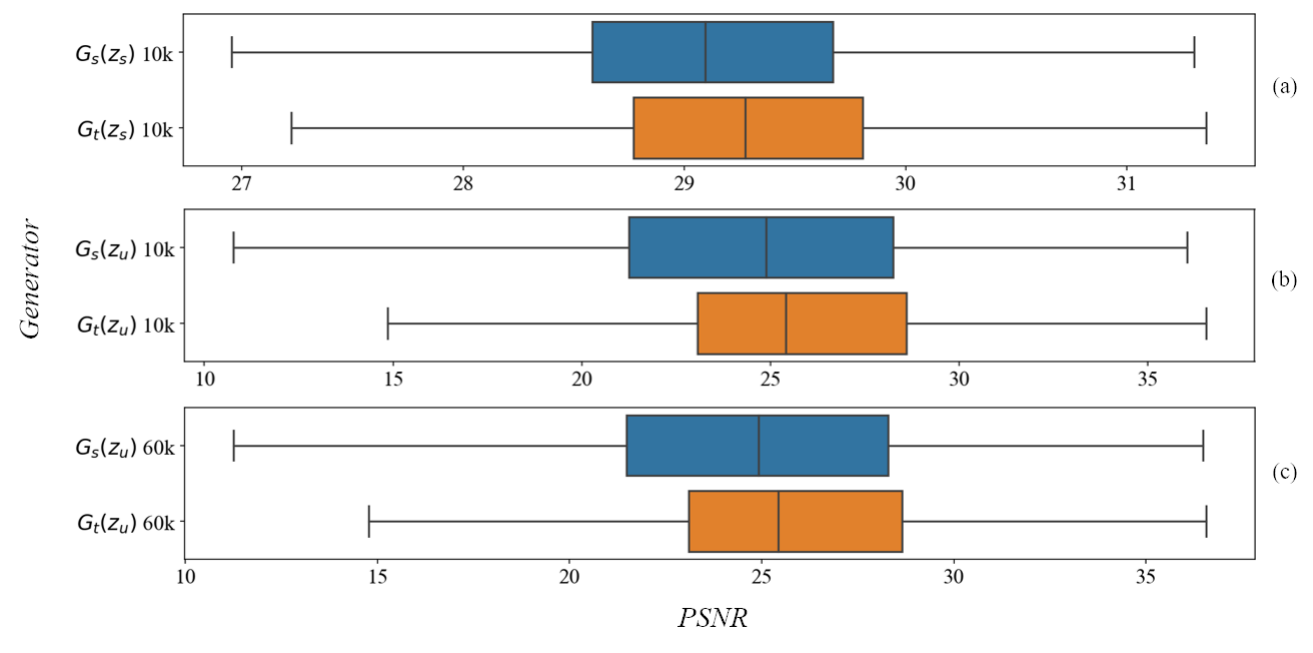}
\caption{Boxplot of PSNR for each generator $G_s$ and $G_t$ on the experimental data set. Even after fine-tuning $G_t$ on unseen data, we observe that $G_t$ produces results without PSNR degradation in the pre-training domain. This finding demonstrates the effectiveness of our proposed method, which utilizes human feedback to mitigate catastrophic forgetting. (a) MNIST test set(10k). (b) Fashion-MNIST test set(10k). (c) Fashion-MNIST train set(60k)}
\label{boxplot_by_generator}
\end{figure}

\noindent {\bf{Adaptive training by human feedback:}} Note that the adaptive training process implements Equation (\ref{final_loss}), utilizing the same set of hyperparameters as mentioned above. It is important to note that $G_t$ has trainable parameters, whereas $\hat{G_s}, \hat{G_t},$ and $\hat{r_\theta}$ are untrainable parameters. The constant $\epsilon$ in Equation (\ref{sigmoid}) is set to 0.2, and the constant $\alpha$ in Equation (\ref{final_loss}) is set to 0.9. In the ablation study, we examine the influence of $L_p$ and $L_{n}$ as $\alpha$ is varied.

\subsection{Evaluation}
We evaluate the quality of the denoised images using the metrics of PSNR (Peak Signal-to-Noise Ratio) and SSIM (Structural Similarity Index Measure). PSNR is a widely used metric for evaluating denoising models. It measures the quality of the denoised image by comparing it to the original (clean) image. Higher PSNR values indicate better denoising performance. PSNR can be calculated using the mean squared error (MSE) between the denoised image and the original image. SSIM is another popular metric that quantifies the similarity between the denoised image and the original image. It takes into account not only pixel-level differences but also structural information, such as luminance, contrast, and structure. Higher SSIM values indicate better preservation of structural details.
\begin{figure*}[!t]
\centering
\subfloat[]{\includegraphics[width=1in]{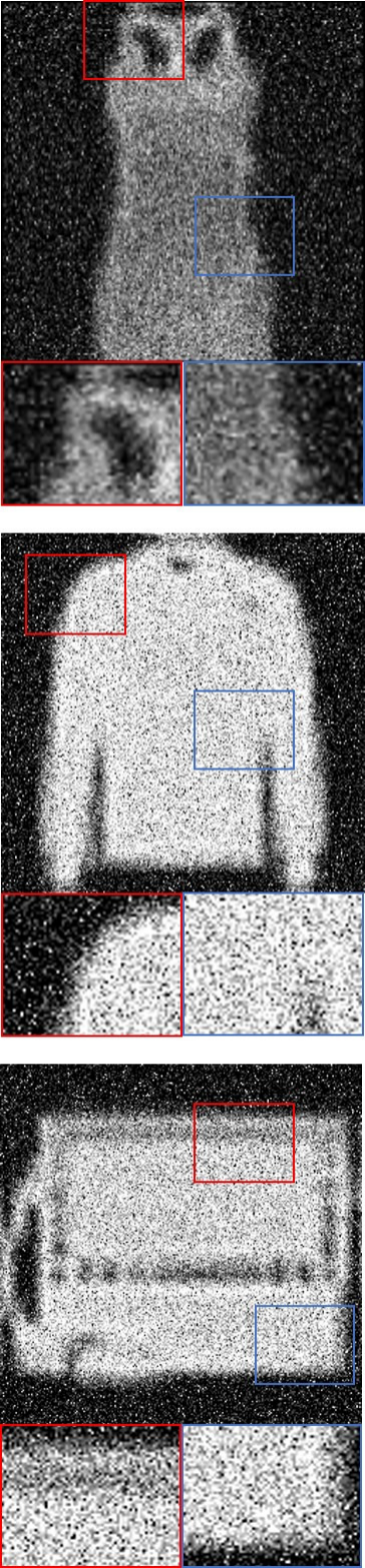}%
\label{cloth_1_input}}
\hfil
\subfloat[]{\includegraphics[width=1in]{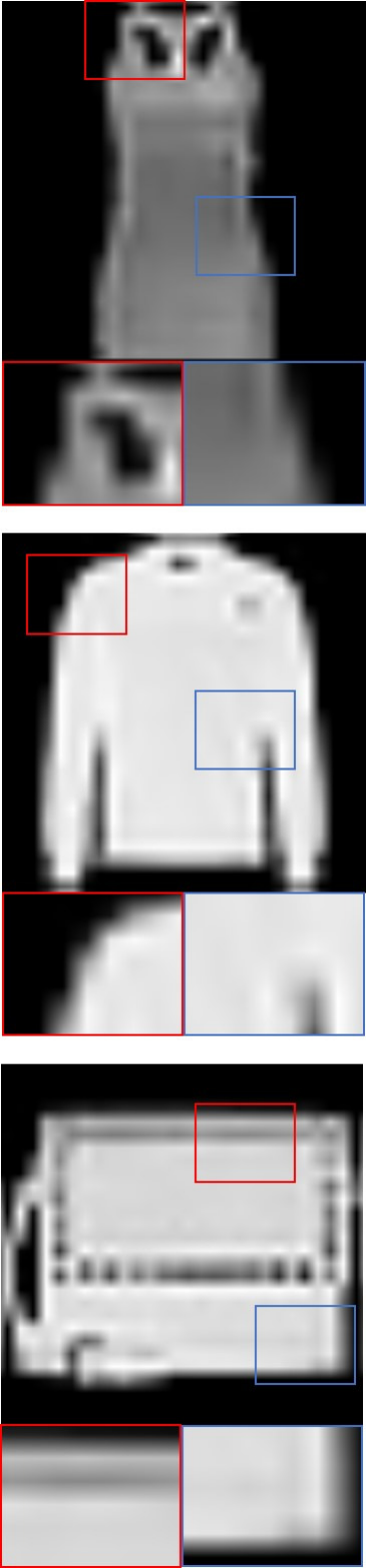}%
\label{cloth_1_ground_truth}}
\hfil
\subfloat[]{\includegraphics[width=1in]{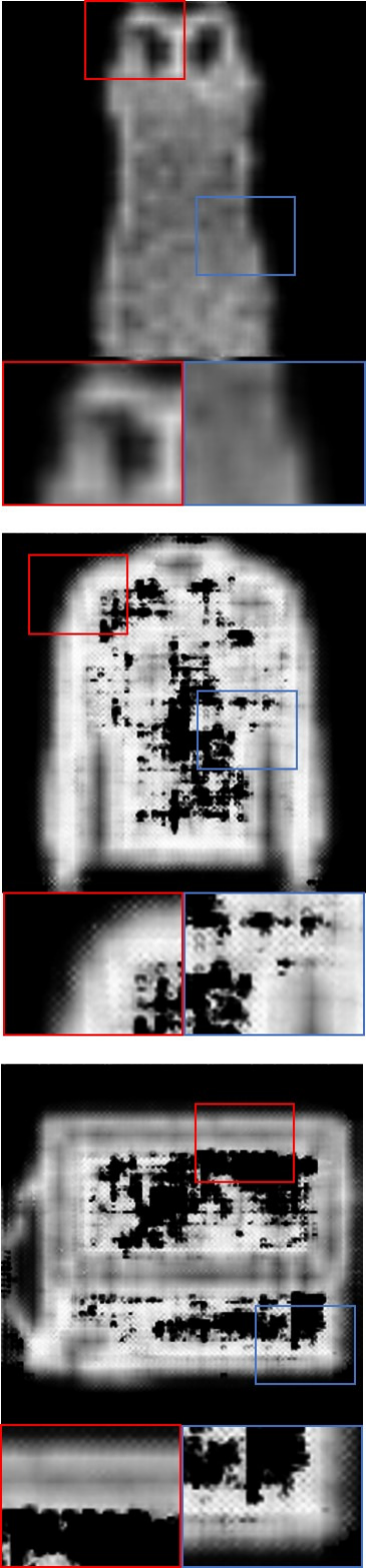}%
\label{cloth_1_G_init}}
\hfil
\subfloat[]{\includegraphics[width=1in]{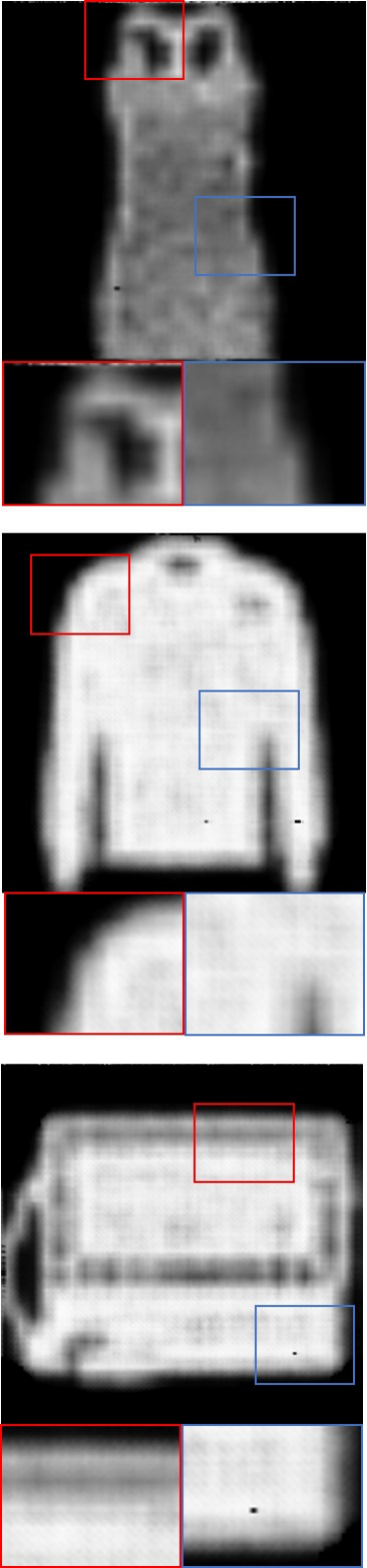}%
\label{cloth_1_G_u}}
\hfil
\subfloat[]{\includegraphics[width=1in]{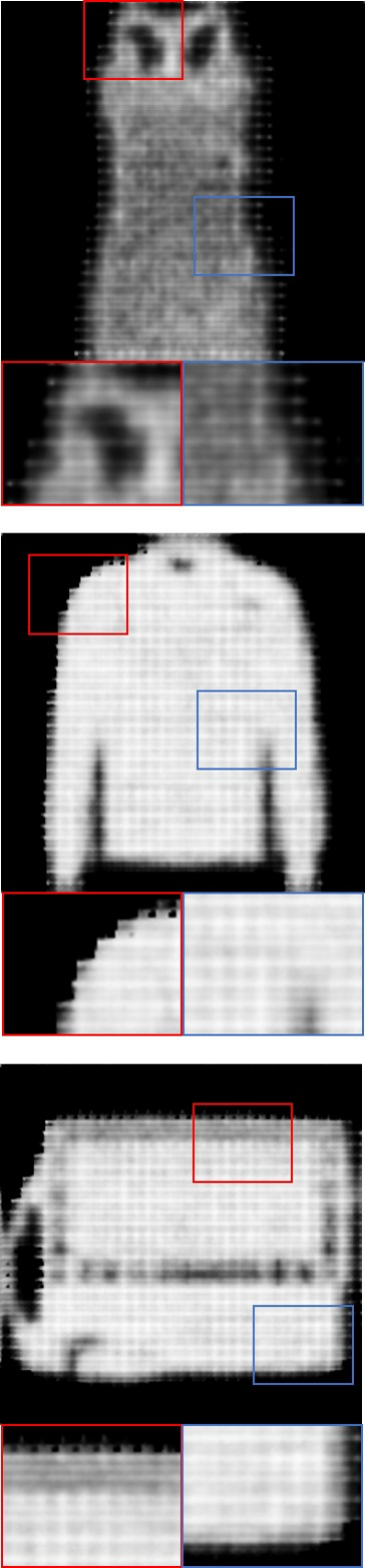}%
\label{cloth_1_no_lp}}
\hfil
\subfloat[]{\includegraphics[width=1in]{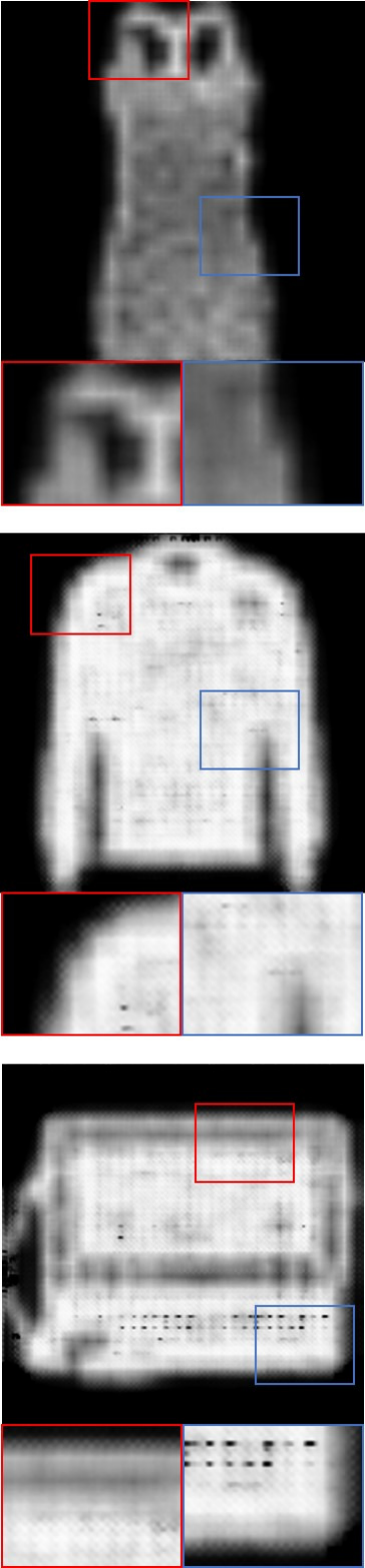}%
\label{cloth_1_no_ln}}
\hfil
\subfloat[]{\includegraphics[width=1in]{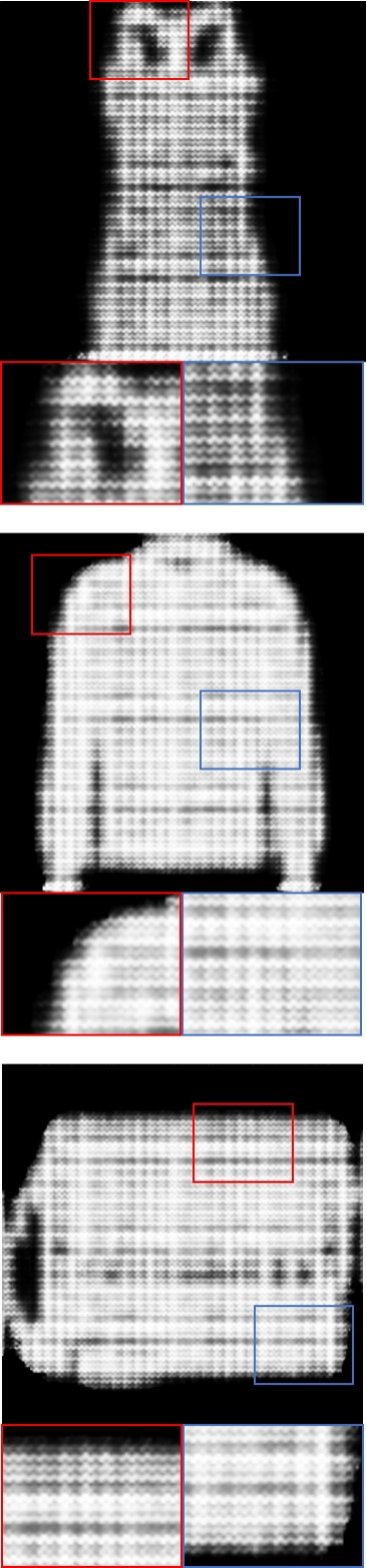}%
\label{cloth_1_only_lr}}
\caption{Comparison of image quality with and without the auxiliary loss. (d), (e), (f), and (g) show results with different auxiliary loss conditions. Each condition improves the image quality compared to (c), but there are noticeable differences in details such as texture and artifacts. (a) Input image with noise. (b) Ground truth. (c) Denoised images by $G_s$. (d) Denoised images by $G_t$. (e) Denoised images by $G_t$ without the $L_p$ term. (f) Denoised images by $G_t$ without the $L_n$ term. (g) Denoised images by $G_t$ using only the $L_r$ term.}
\label{fig_sim}
\end{figure*}
\subsection{Results of domain adaptive denoising by human feedback}
\noindent {\bf{Comparison of evaluation metrics:}} In this section, we examine the results of domain adaptive denoising. Our intuition is that, even when presented with unseen data from a target domain, if we provide human feedback to a supervised learning model, the model can adapt to the data effectively. Note that our human feedback is not ground-truth for denoised image, it is human's preference consisting of `$Good$' and `$Bad$'. In Table \ref{result_FT_HF}, `$G_s(z)$ vs $x$' represents the denoising results of the model before adaptive learning using human feedback. `$G_t(z)$ vs $x$' shows the denoising outcomes of the adapted model based on human feedback. The results obtained from the MNIST data set indicate the performance in the pre-trained domain, while the results from the Fashion-MNIST data set reflect the performance on unseen data. Therefore, we can observe the adaptation progress between the initial model $G_s$ and the updated model $G_t$. In our experiments, both $G_s$ and $G_t$ demonstrated a significant improvement in PSNR measured on the Fashion-MNIST test set, with an overall increase of 94\% over the entire 10K data set. The statistical analysis of the PSNR improvement revealed a mean increase of 1.61$\pm$2.78$dB$ (MAX: 18.21, MIN: 0.0001). Fig. \ref{visual_result_for_denoising}-(a) shows the images with the most significant increase in PSNR, along with the corresponding metrics between the each generator output images and the ground truth images. In addition, for the remaining 6\% of the cases, there was a decrease in PSNR values, with the statistical analysis showing a mean decrease of 0.12$\pm$0.18$dB$ (MAX: 2.75, MIN: 0.0001)(See. Fig. \ref{visual_result_for_denoising}-(b)). Fig. \ref{boxplot_by_generator} shows the boxplot of PSNR for each generator $G_s$ and $G_t$ on the experimental data set. The $G_t$ images from the same data set exhibit higher PSNR values, indicating improved image quality after adaptation. Particularly noteworthy is that after adaptation, the PSNR and SSIM values of the MNIST test set(10K) from the $G_t$ generator, corresponding to the source domain, show little to no variation or even slight improvement(See. Fig \ref{boxplot_by_generator}-(a)). This demonstrates the prevention of catastrophic forgetting issue for the source domain even after adaptation to the target domain. Furthermore, we apply the $G_t$ model tuned on the Fashion-MNIST test set with reward model to the Fashion-MNIST training set (60k). This demonstrates that when the reward model is trained in a new domain, it can effectively work without requiring additional training.\\

\noindent {\bf{Visual evaluation:}} Fig. \ref{visual_result_for_denoising} illustrates the improvement in denoising and restoration, particularly in addressing image collapse. Notably, $G_s$ trained on `0' digit of MNIST, exhibits instances where the results suffer from image collapse in several images, indicating a lack of adaptation. However, our approach effectively enhances the image quality by leveraging human feedback, as demonstrated by the results obtained with $G_t$.

\subsection{Ablation study}
To  validate  the  effectiveness  of  each  loss term  of  our  method. we conduct comprehensive ablation studies for loss term
%we conduct experiments ablation study for loss term.

\noindent {\bf{Effect of $L_p$ term:}} The $Lp$ term compares the image quality between $G_s$ and $G_t$, and is the loss function between images that are well evaluated by human feedback based on the reward function. We examine the effect of the $L_p$ loss on the quality of the output. Typically, the constant alpha of $L_p$ is fixed at 0.9. To evaluate the effect of excluding the $L_p$ term, we vary the alpha value to 0, resulting in the loss equation becoming $L(G_t) = L_r(G_t) + L_{n}(G_t)$. Performing the adaptation without an $L_p$ term exhibits low quantitative performance, as demonstrated in the second row of Table \ref{result_FT_HF}. Additionally, Fig. \ref{fig_sim}(d)-(e) depict the anomaly texture created image for reference.\\
% ksh 
\noindent {\bf{Effect of $L_n$ term:}} $L_n$ represents the $L_1$ loss between the ($n-2$)th and ($n$)th iterations of $G$($\hat{G_t}$ and $G_t$). 
In terms of quantitative evaluation, it demonstrates comparable performance (Table 1, first and third row). However, in qualitative assessment, it becomes evident that there are limitations in generating the desired image to a satisfactory degree.(See Fig. \ref{fig_sim}(d)-(f)).
We also examine the effect of $L_n$ loss on the output quality. The role of $L_n$ is to restrict significant parameter changes from the previous model.
Given that the function of $L_n$ is to restrict parameter updates between the previous and current models, it becomes apparent that there are limitations in generating the desired image to a satisfactory extent in qualitative evaluation, leading to potential issues such as collapse.\\

\noindent {\bf{Effect of $L_r$ term:}} $L_r$ represents .
In the experiments where $L_p$ and $L_n$ are ablated, the adaptation relies solely on $L_r$, resulting in parameter updates exclusively driven by human feedback. Consequently, in the absence of pixel-wise losses such as $L_p$ and $L_n$, it is evident that image details and shapes are not preserved, as illustrated in Fig. \ref{fig_sim} (d)-(g).

\section{Discussion and Conclusions}
In this paper, we propose a novel method based on human feedback to address the domain adaptation problem of denoising generative models, particularly focusing on the condition of an unlabeled target domain. Unlike conventional approaches that aim to enhance a generative model's overall performance on the entire test data set, our method leverages human feedback to directly improve the quality of failed images in denoising tasks. While many existing approaches require a large amount of labeled data and may discard failed images, our approach fine-tunes the model based on human feedback, similar to the process used in ChatGPT\cite{brown2020language} to select generated sentences of `Good' quality. This novel utilization of human feedback represents a promising avenue for enhancing generative models. 

Domain adaptation poses challenges, particularly regarding the issue of catastrophic forgetting. However, through our proposed adaptation approach, which incorporates selective loss functions and an ablation study based on decisions from a reward model trained with human feedback, we successfully mitigated the challenging issue of catastrophic forgetting. Our results align with related studies, demonstrating the effectiveness of our approach.

In the context of real-world applications, the unseen data domain adaptation of deep generative models has always been a crucial research topic. In this paper, we demonstrate the adaptation of a model trained on the source domain to the label-less target domain, guided by human feedback. Through ablation study, we analyzed the loss functions and provided compelling evidence for the direction of domain adaptation research, particularly in the realm of image generation. Furthermore, we will grapple for two things as follows:
1. Human preference: Our work also collect human feedback data by personal preference same with ChatGPT. Thus, distribution for `Good' quality can be different. This will be connected directly with model's performance.
2. Model performance dependent on pre-training: We assume that SDM is over a certain level. However, if SDM does not work in unseen domain not at all, we can not collect human feedback. Human feedback has to be collected `Good' and `Bad' categroy.

\section*{Acknowledgments}
This work was supported by the National Institute for Mathematical Sciences (NIMS) funded by the Korean Government under Grant NIMS-B23910000.

%%
%\bibliographystyle{IEEEtran}
%% argument is your BibTeX string definitions and bibliography database(s)
%\bibliography{IEEEabrv, manuscript}


\newpage

\vfill

\end{document}